\documentclass[letterpaper, 10 pt, conference]{ieeeconf}  

\IEEEoverridecommandlockouts                              

\usepackage{times}
\usepackage{soul}
\usepackage{url}
\usepackage[hidelinks]{hyperref}
\usepackage[utf8]{inputenc}
\usepackage[small]{caption}
\usepackage{graphicx}
\usepackage{amsmath}
\usepackage{amsthm}
\usepackage{booktabs}
\usepackage{algorithm}
\usepackage{algorithmic}
\usepackage[switch]{lineno}
\usepackage{graphics} 
\usepackage{graphicx}
\usepackage{amssymb}  
\usepackage{cleveref}
\usepackage{caption}
\usepackage{siunitx}
\usepackage[font=footnotesize]{caption}
\usepackage{float}
\usepackage{subfig}
\usepackage[dvipsnames]{xcolor}
\usepackage{color, colortbl}

\title{\LARGE \bf
Learning control of underactuated double pendulum with Model-Based Reinforcement Learning
}

\author{Niccolò Turcato$^1$, Alberto Dalla Libera$^1$, Giulio Giacomuzzo$^1$, Ruggero Carli$^1$ and Diego Romeres$^2$
\thanks{}
\thanks{$^{1}$Department of Information Engineering, University of Padova, Italy.}%
\thanks{$^{2}$Mitsubishi Electric Research Laboratories, Cambridge, MA, USA}%
\thanks{Correspondence to {\tt\small niccolo.turcato@phd.unipd.it}}%
}

\begin{document}

\definecolor{LightRed}{rgb}{1,0.7,0.7}
\newcommand{\R}{\mathbb{R}}
\newcommand{\q}{\boldsymbol{q}}
\newcommand{\dq}{\boldsymbol{\dot{q}}}
\newcommand{\ddq}{\boldsymbol{\ddot{q}}}
\newcommand{\taubf}{\boldsymbol{\tau}}
\newcommand{\cor}{\boldsymbol{c}}
\newcommand{\X}{\boldsymbol{X}}
\newcommand{\x}{\boldsymbol{x}}
\newcommand{\ub}{\boldsymbol{u}}
\newcommand{\ab}{\boldsymbol{a}}
\newcommand{\thetab}{\boldsymbol{\theta}}
\newcommand{\Y}{\boldsymbol{Y}}
\newcommand{\y}{\boldsymbol{y}}
\newcommand{\w}{\boldsymbol{w}}
\newcommand{\p}{\boldsymbol{p}}
\newcommand{\Pb}{\boldsymbol{P}}
\newcommand{\Deltab}{\boldsymbol{\Delta}}
\newcommand{\deltab}{\boldsymbol{\delta}}
\newcommand{\Lag}{\mathcal{L}}
\newcommand{\E}{\mathbb{E}}
\newcommand{\norm}[1]{\left\lVert#1\right\rVert}
\newcommand{\D}{\mathcal{D}}
\newcommand{\f}{\mathbf{f}}
\newcommand{\e}{\boldsymbol{e}}
\newcommand{\mub}{\boldsymbol{\mu}}
\newcommand{\Fb}{\boldsymbol{F}}
\newcommand{\TODO}{\textcolor{red}{\textbf{TODO}}}

\maketitle
\thispagestyle{empty}
\pagestyle{empty}

\begin{abstract}
This report describes our proposed solution for the second AI Olympics competition held at IROS 2024. Our solution is based on a recent Model-Based Reinforcement Learning algorithm named MC-PILCO. Besides briefly reviewing the algorithm, we discuss the most critical aspects of the MC-PILCO implementation in the tasks at hand. 

\end{abstract}


\section{Introduction}
In this short report, we present the Reinforcement Learning (RL) \cite{sutton2018reinforcement_learning} approach our team implemented to tackle the simulation stage of the second AI Olympics competition held at IROS 2024\footnote{\url{https://ai-olympics.dfki-bremen.de/}}. The algorithm we employed, Monte-Carlo Probabilistic Inference for Learning COntrol (MC-PILCO) \cite{amadio2021mc_pilco}, is a Model-Based (MB) RL algorithm that proved remarkably data-efficient in several low-dimensional benchmarks, such as a cart-pole, a ball \& plate, and a Furuta pendulum, both in simulation and real setups. MC-PILCO is also the algorithm that won the first edition of this competition \cite{ijcai2024p1043}.
MC-PILCO is part of the class of MB policy gradient algorithms. It exploits data collected by interacting with the system to derive a system dynamics model and optimizes the policy by simulating the system, rather than optimizing the policy directly on the system's data.
When applied to physical systems, this approach can be highly performing and more data-efficient than Model-Free (MF) solutions.
Examples of MC-PILCO applications have been reported in \cite{amadio2023mcpilco_raw_meas} and \cite{mcpilco_tossing}. 

This paper is organized as follows: \Cref{sec:goal} introduces the goal and the settings of the competition. \Cref{sec:methods} presents the MC-PILCO algorithm. \Cref{sec:experiments} reports the experiments that have been performed, finally \Cref{sec:conclusions} concludes the paper.

\section{Goal of the competition}
\label{sec:goal}
The challenge considers a 2 degrees of freedom (dof) underactuated pendulum \cite{10375556} with two possible configurations. In the first configuration, also called Pendubot, the first joint, namely, the one attached to the base link is active, and the second is passive. Instead, in the second configuration, also named Acrobot, the first joint is passive and the second is actuated. For each configuration, the competition's goal is to derive a controller that performs the swing-up and stabilization in the unstable equilibrium point of the systems. Both robots are underactuated, which makes the task particularly challenging from the control point of view. The systems are simulated at \SI{500}{\hertz} with a Runge-Kutta 4 integrator for a horizon of $T=\SI{10}{\s}$. The competition is composed of 2 stages. In the first stage, namely the simulation stage, controllers are assessed based on performance and robustness scores in the simulated system. In the second stage, namely the real hardware stage, the participating teams test their controllers on the real system, with the possibility of retraining their learning-based controllers. The winners of the competition are chosen based on the performance and the reliability of the submitted controllers.

\section{MC-PILCO for underactuated robotics}
\label{sec:methods}
In this section, firstly we review MC-PILCO, secondly, we discuss its application to the considered problem. 

\subsection{MC-PILCO review}
\label{subsec:mc_pilco_review}
MC-PILCO is a MB policy gradient algorithm, in which GPs are used to estimate system dynamics and long-term state distributions are approximated with a particle-based method.

Consider a system with evolution described by the discrete-time unknown transition function $f: \R^{d_x} \times \R^{d_u} \rightarrow \R^{d_x}$:
\begin{equation}
    \x_{t+1} = f(\x_{t}, \ub_{t}) + \w_{t},
    \label{eq:system_equation}
\end{equation}
where $\x_{t} \in \R^{d_x}$ and $\ub_{t} \in \R^{d_u}$ are respectively the state and input of the system at step $t$, while $\w_{t}$ is an independent white noise describing uncertainty influencing the system evolution. As usual in RL, a cost function $c(\x_{t})$ encodes the task to be accomplished. A policy $\pi_{\thetab}: \x \rightarrow \ub$ that depends on the parameters $\thetab$ selects the inputs applied to the system. The objective is to find policy parameters $\thetab^*$ that minimize the cumulative expected cost, defined as follows,
\begin{equation}
    J(\thetab) = \sum_{t=0}^T \E [c(\x_{t})],
    \label{eq:cumulative_cost}
\end{equation}
where the initial state $x_0$ is sampled according to a given probability $p(\x_0)$.

MC-PILCO consists of a series of attempts, known as trials, to solve the desired task.
Each trial consists of three main phases: (i) model learning, (ii) policy update, and (iii) policy execution. In the first trial, the GP model is derived from data collected with an exploration policy, for instance, a random exploration policy. 

In the model learning step, previous experience is used to build or update a model of the system dynamics. The policy update step formulates an optimization problem whose objective is to minimize the cost in \cref{eq:cumulative_cost} w.r.t. the parameters of the policy $\thetab$. 
Finally, in the last step, the current optimized policy is applied to the system and the collected samples are stored to update the model in the next trials.

In the rest of this section, we give a brief overview of the main components of the algorithm and highlight their most relevant features.

\subsubsection{Model Learning}
\label{subsubsec:model_learning}
MC-PILCO relies on GP Regression (GPR) to learn the system dynamics \cite{rasmussen2003gps_for_ml}. For the use of GPs in system identification and control we refer the interested reader to
\cite{kernel_methods_and_gp_control_systems_magazine}.
In our previous work, \cite{amadio2021mc_pilco}, we presented a framework specifically designed for mechanical systems, named speed-integration model. 
Given a mechanical system with $d$ degrees of freedom, the state is defined as $\x_t = [\q_t^T, \dq_t^T]^T$ where $\q_t \in \mathbb{R}^d$ and $\dq_t \in \mathbb{R}^d$ are, respectively, the generalized positions and velocities of the system at time $t$. 
Let $T_s$ be the sampling time and assume that accelerations between successive time steps are constant. 
The following equations describe the one-step-ahead evolution of the $i$-th degree of freedom, 

\begin{subequations} \label{eq:speed-int}
\begin{align}
    &\dot{q}^{(i)}_{t+1} = \dot{q}^{(i)}_{t} + \Delta ^{(i)}_t \label{eq:speed-int-vel}\\
    &q_{t + 1}^{(i)} = q_t^{(i)} + T_s \dot{q}^{(i)}_t + \frac{T_s}{2} \Delta ^{(i)}_t \label{eq:speed-pos}
\end{align}
\end{subequations}
where $\Delta ^{(i)}_t$ is the change in velocity. MC-PILCO estimates the unknown function $\Delta^{(i)}_t$ from collected data by GPR. Each $\Delta_t^{(i)}$ is modeled as an independent GP, denoted $f^i$, with input vector $\tilde{\x}_t = [\x^T_t, \ub^T_t]^T$, hereafter referred as GP input. Given an input-output training dataset $\D^{(i)} = \{ \tilde{\X}, \y^{(i)}\}$, where the inputs are $\tilde{\X} = [\tilde{\x}_1^T, \dots, \tilde{\x}_n^T]^T$, and the outputs $\y^{(i)}=[y^{(i)}_1, \dots y^{(i)}_n]^T$ are measurements of $\Delta ^{(i)}_t$ at time instants $t = 0, \dots, T_{tr}$, GPR assumes the following probabilistic model,
\begin{equation}
    \y^{(i)} = f^i(\tilde{\X}) + \e,
\end{equation}
where vector $\e$ accounts for noise, defined a priori as zero mean independent Gaussian noise with variance $\sigma_i^2$. The unknown function $f^i$ is defined a priori as a GP with mean $m^{(i)}_\Delta$ and covariance defined by a kernel function $k(\tilde{\x}_{t_i},\tilde{\x}_{t_j})$, namely,  $f^i(\tilde{\X}) \sim N(m^{(i)}_\Delta, K_{\tilde{\X} \tilde{\X}})$, where the element of $K_{\tilde{\X} \tilde{\X}}$ at row $r$ and column $j$ is $E[\Delta ^{(i)}_{t_r}, \Delta ^{(i)}_{t_j}] = k(\tilde{\x}_{t_r},\tilde{\x}_{t_j})$. The mean function $m^{(i)}_\Delta$ can be derived from prior knowledge of the system, or can be set as the null function if no information is available. Instead, as regards the kernel function, one typical choice to model continuous functions is the squared-exponential kernel: 
\begin{equation}\label{eq:se-kernel}
    k(\tilde{\x}_{t_i}, \tilde{\x}_{t_j}):= \lambda^2 \e^{-\norm{\tilde{\x}_{t_i}- \tilde{\x}_{t_j}}^2_{\Lambda^{-1}}}
\end{equation}
where $\lambda$ and $\Lambda$ are trainable hyperparameters tunable by maximizing the marginal likelihood (ML) of the training samples \cite{rasmussen2003gps_for_ml}.

As explained in \cite{rasmussen2003gps_for_ml}, the posterior distributions of each $\Delta_t^{(i)}$ given $\mathcal{D}^{i}$ are Gaussian distributed, with mean and variance expressed as follows:
\begin{equation}
    \begin{split}
        &\E[\hat{\Delta}_t^{(i)}] = m^{(i)}_\Delta(\tilde{\x}_t) + K_{\tilde{\x}_t \tilde{\X}} \Gamma_i^{-1} (\y^{(i)} - m^{(i)}_\Delta(\tilde{\X}))    \\
        &var[\hat{\Delta}_t^{(i)}] = k_i(\tilde{\x}_t, \tilde{\x}_t) - K_{\tilde{\x}_t \tilde{\X}} \Gamma_i^{-1} K_{\tilde{\X} \tilde{\x}_t} \\
        &\Gamma_i = K_{\tilde{\X} \tilde{\X}} + \sigma_i^2 I
    \end{split}
    \label{eq:gp_regression_formulas}
\end{equation}
Then, also the posterior distribution of the one-step ahead transition model in \eqref{eq:speed-int} is Gaussian, namely, 
\begin{equation}
    p(\x_{t+1} | \x_t, \ub_t, \D) \sim \mathcal{N}(\mub_{t+1}, \Sigma_{t+1}) 
    \label{eq:one-step-posteior}
\end{equation}
with mean $\mub_{t+1}$ and covariance $\Sigma_{t+1}$ derived combining \eqref{eq:speed-int} and \eqref{eq:gp_regression_formulas}.

\subsubsection{Policy Update}
\label{subsubsec:policy_update}
In the policy update phase, the policy is trained to minimize the expected cumulative cost in \cref{eq:cumulative_cost} with the expectation computed w.r.t. the one-step ahead probabilistic model in \cref{eq:one-step-posteior}. This requires the computation of long-term distributions starting from the initial distribution $p(\x_0)$ and \cref{eq:one-step-posteior}, which is not possible in closed form. MC-PILCO resorts to Monte Carlo sampling \cite{caflisch1998monte_carlo_sampling_ref} to approximate the expectation in \cref{eq:cumulative_cost}. The Monte Carlo procedure starts by sampling from $p(\x_0)$ a batch of $N$ particles and simulates their evolution based on the one-step-ahead evolution in \cref{eq:one-step-posteior} and the current policy. Then, the expectations in \cref{eq:cumulative_cost} are approximated by the mean of the simulated particles costs, namely,

\begin{equation}
    \begin{split}
        \hat{J}(\boldsymbol{\thetab}) = \sum_{t=0}^{T} \left( \frac{1}{N} \sum_{n=1}^N c \left( \x_t^{(n)} \right) \right) 
    \end{split}
    \label{eq:cost_estimate_monte_carlo}
\end{equation}
where $\x_t^{(n)}$ is the state of the $n$-th particle at time $t$.

The optimization problem is interpreted as a stochastic gradient descend problem (SGD) \cite{bottou2010large_scale_learning_sgd}, applying the reparameterization trick to differentiate stochastic operations \cite{kingma2013reparametrization_trick}.

The authors of \cite{amadio2021mc_pilco} proposed the use of dropout \cite{srivastava2014dropout} of the policy parameters $\thetab$ to improve exploration and increase the ability to escape from local minima during policy optimization of MC-PILCO.

\subsection{MC-PILCO for underactuated robotics}
\label{subsec:mcpilco_underact}
The task in object presents several practical issues when applying the algorithm. The first one is that the control frequency requested by the challenge is quite high for a MBRL approach. Indeed, high control frequencies require a high number of model evaluations which increases the computational cost of the algorithm. Generally, this class of systems can be controlled at relatively low frequencies, for instance, \cite{amadio2021mc_pilco} and \cite{amadio2023learning} derived a MBRL controller for a Furuta Pendulum at \SI{33}{\hertz}. Indeed, in the real hardware stage of the first edition of the competition, the MC-PILCO controller was trained to work at \SI{33}{\hertz}.
However, the physical properties of the simulated systems (no friction) make the system particularly sensitive to the system input. For these reasons, we selected a control frequency of \SI{50}{\hertz}.

The second issue is that controllers are evaluated by a performance and robustness score. In the robustness test, the characteristics of the system and data acquisition vary. This is an issue for data-driven solutions like MC-PILCO since retraining of the controller is not allowed. For this reason, we decided to focus only on solving the swingup task on the nominal system, even if in our previous work we showed that MC-PILCO can be robust to noise and filtering by including these effects in the simulation.

Since the nominal model of the system is available to develop the controller, we use the forward dynamics function of the plant as the prior mean function of the change in velocity for each joint. The available model is
\begin{equation}
    B \ub_t = M(\q_t) \ddq_t + n(\q_t,\dq_t),
\end{equation}
where $M(\q_t)$ is the mass matrix, $n(\q_t,\dq_t)$ contains the Coriolis, gravitational and damping terms, and $B$ is the actuation matrix, which is $B=\text{diag}([1, 0])$ for the Pendubot and $B=\text{diag}([0, 1])$ for the Acrobot.
We define then
\begin{equation}
    m_\Delta(\tilde{\x}_t) = \begin{bmatrix} m_\Delta^{(1)} \\ m_\Delta^{(2)}\end{bmatrix} := T_s \cdot M^{-1}(\q_t) (B \ub_t - n(\q_t,\dq_t))
    \label{eq:mean_fun_forward_dyn}
\end{equation}
as the mean function in \cref{eq:gp_regression_formulas}.
It is important to point out that \cref{eq:mean_fun_forward_dyn} is nearly perfect to approximate the system when $T_s$ is sufficiently small, but it becomes unreliable as $T_s$ grows. In particular, with $T_s=\SI{0.02}{\s}$ the predictions of \cref{eq:mean_fun_forward_dyn} are not good enough to describe the behavior at the unstable equilibrium. The inaccuracies of the prior mean are compensated by the GP models. To cope with the large computational burden due to the high number of collected samples, we implemented the GP approximation Subset of Regressors, see \cite{JMLR:v6:quinonero-candela05a} for a detailed description. 

An important aspect of policy optimization is the particles initialization, in this case, it is guaranteed that the system will always start at $\x_0=\Bar{0}$, therefore the initial distribution can be set to $p(\x_0) \sim \mathcal{N} (\Bar{0}, \epsilon  I)$ with $\epsilon$ in the order of $10^{-4}$.

The cost function must evaluate the policy performance w.r.t. the task requirements, in this case, we want the system to reach the position $\q_G=[\pi, 0]^T$ and stay there indefinitely. A cost generally used in this kind of system is the saturated distance from the target state:
\begin{align}
    \begin{split}
        c_{st}(\x_t) = 1 - e^{- \norm{\q_t - \q_G}^2_{\Sigma_c}}   \hspace{0.5cm}
        \Sigma_c = \text{diag}\left(\frac{1}{\ell_c}, \frac{1}{\ell_c}\right),
    \end{split}
    \label{eq:saturated_dist_target}
\end{align}
with $\ell_c=3$. Notice that this cost does not depend on the velocity of the system, just on the distance from the goal state, but it does encourage the policy to reach the goal state with zero velocity.

The policy function that is used to learn a control strategy is the general purpose policy from \cite{amadio2021mc_pilco}:
\begin{equation}
    \begin{split}
         &\pi_{\thetab}(\x_t) = u_M \tanh{ \left(  \sum_{i=1}^{N_b} \frac{w_i}{u_{M}} e^{-\| \ab_i - \phi(\x_t) \|^2_{\Sigma_\pi}}  \right)} \\
         &\phi(\x_t) = [\dq_t^T, \cos{(\q_t^T)}, \sin{(\q_t^T)}]^T\\
    \end{split}
    \label{eq:tossing_policy}
\end{equation}

with hyperparameters $\thetab = \{{\bf w}, A, \Sigma_\pi\}$, where ${\bf w} = [w_1, \dots, w_{N_b}]^T$ and $A= \{\ab_1, \dots, \ab_{N_b}\}$ are, respectively, weights and centers of the $N_b$ Gaussians basis functions, whose shapes are determined by $\Sigma_{\pi}$. For both robots, the dimensions of the elements of the policy are: $\Sigma_{\pi} \in \R^{6 \times 6}$, $a_i \in \R^6$, $w_i \in \R$ for $i = 1, \dots, N_b$, since the policy outputs a single scalar. In the experiments, the parameters are initialized as follows. The basis weights are sampled uniformly in $[-u_M, u_M]$, the centers are sampled uniformly in the image of $\phi$ with $\dq_t \in [-2 \pi, 2 \pi]$ rad/s. The matrix $\Sigma_{\pi}$ is initialized to the identity. Given the ideal conditions considered in this simulation, for the purpose of the challenge, the control switches to an LQR controller after the swing-up.
Under ideal circumstances, the LQR controller has the capability to stabilize the system at an unstable equilibrium by exerting zero final torque. The switching condition is obtained by checking if the system's state is within the LQR's region of attraction.

\section{Experiments}
\label{sec:experiments}
In this section, we briefly discuss how the algorithm was applied to both systems and show the main results. We also report the optimization parameters used for both systems, all the parameters not specified are set to the values reported in \cite{amadio2021mc_pilco}.
All the code was implemented in Python with the PyTorch \cite{paszke2017automatic_diff_pytorch} library.

For both robots, we use the model described in \Cref{subsubsec:model_learning}, with mean function from \cref{eq:mean_fun_forward_dyn} and kernel function from \cref{eq:se-kernel}.
The max torque $u_{M}$ was set to conservative values, to improve the performance score of the controller. The policy optimization horizon was set much lower than the horizon required for the competition, this allows to reduce the computational burden of the algorithm, moreover, it pushes the optimization to find policies that can execute a fast swing-up. We exploit dropout in the policy optimization as a regularization strategy, to yield better policies.

\subsection{Pendubot}
\label{sec:pendubot}

\begin{figure}
    \centering	
    \includegraphics[width=\columnwidth]{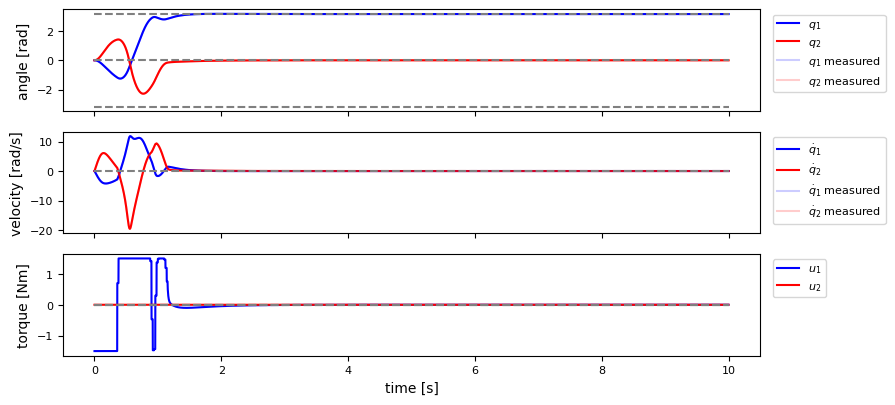}
    \caption{Simulation of the Pendubot system (\SI{500}{\hertz}), under control of the policy trained with MC-PILCO.}
    \label{fig:pendubot_rollout}
\end{figure}

The policy for the Pendubot swing-up was optimized for a horizon of $T=\SI{3.0}{\s}$, with $u_M$ set to $25$\% of the torque limit of the actuator. 
The Controller's strategy is depicted in \cref{fig:pendubot_rollout}, in \cref{fig:robust_bar_charts} (left) we report the robustness bar charts. This controller has a performance score of $0.48$ and a robustness score of $0.61$.
In \cref{tab:pendubot_table} we compare our controller's score with other tested control strategies.
\begin{center}
\begin{table}[H]

\begin{tabular}{||c c c c||} 
 \hline
 Controller & Perf. score & Rob. score & Avg. score \\ [0.5ex] 
 \hline \hline
TVLQR & 0.526 & 0.767 & 0,647\\
\hline
\rowcolor{LightRed} MC-PILCO & 0.48 & 0.61 & 0.545 \\
 \hline
iLQR MPC stab. & 0.353 & 0.674 & 0.514 \\
\hline 
iLQR Riccati & 0.536 & 0.255 & 0.396 \\
\hline

\end{tabular}
\caption{Penubot scores comparison.}
\label{tab:pendubot_table}
\end{table}
\end{center}

\subsection{Acrobot}
\label{sec:acrobot}
\begin{figure}
    \centering	
    \includegraphics[width=\columnwidth]{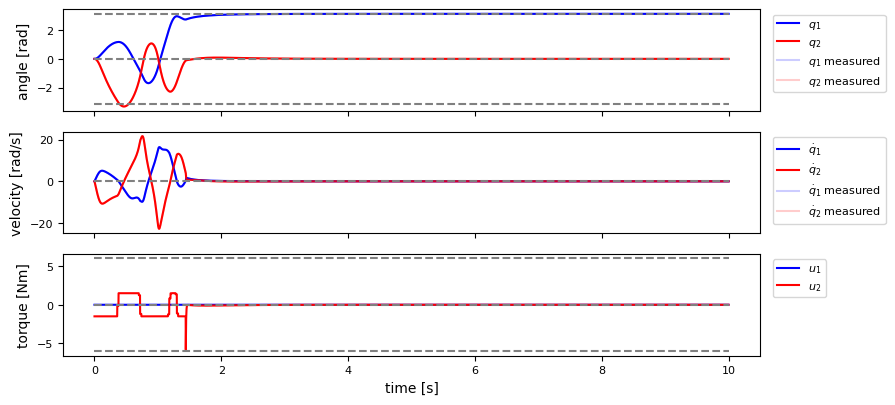}
    \caption{Simulation of the Acrobot system (\SI{500}{\hertz}), under control of the policy trained with MC-PILCO.}
    \label{fig:acrobot_rollout}
\end{figure}

\begin{figure}[H]
\footnotesize
    \subfloat{%
      \includegraphics[width=0.45\columnwidth]{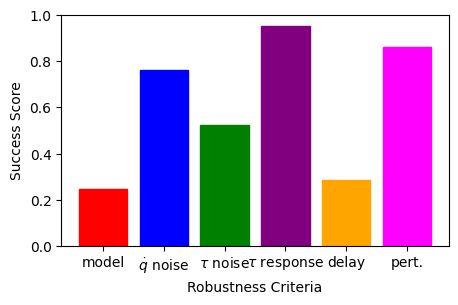}%
    }
    \hfill
    \subfloat{%
      \includegraphics[width=0.45\columnwidth]{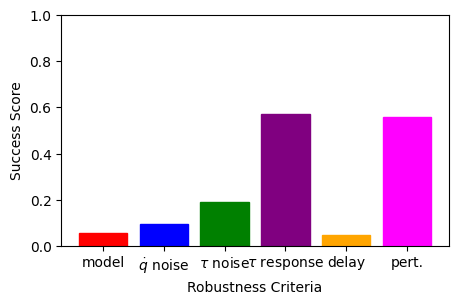}%
    }
    \caption{Pendubot (left) and Acrobot (right) robustness bar charts.}
    \label{fig:robust_bar_charts}
\end{figure}

The policy for the Acrobot swing-up was optimized for a horizon of $T=\SI{2.0}{\s}$, with $u_M$ set to $25$\% of the torque limit of the actuator. 
The Controller's strategy is depicted in \cref{fig:acrobot_rollout}, in \cref{fig:robust_bar_charts} (right) we report the robustness bar charts. This controller has a performance score of $0.316$ and a robustness score of $0.25$. 
In \cref{tab:acrobot_table} we compare our controller's score with other tested control strategies.

\section{Conclusions}
\label{sec:conclusions}
In both systems, our MBRL approach is able to solve the task with very good swing-up time, comparable with the result of the first edition. The performance scores of the controllers are not very competitive, w.r.t. the baselines in the leaderboard\footnote{\url{https://dfki-ric-underactuated-lab.github.io/real_ai_gym_leaderboard/}}, since the score penalizes energy consumption, velocity and torque smoothness, which are not penalized in the cost function in \cref{eq:saturated_dist_target}. As already seen in the first edition, MC-PILCO controllers are sensitive to parameter changes and delays, since they are not present in the data seen in training. However, the possibility of retraining on real hardware with few trials is a great strength of this approach.
Lastly, the controllers proved a certain level of robustness when subject to actuation perturbations.

\begin{center}
\begin{table}[H]

\begin{tabular}{||c c c c||} 
 \hline
 Controller & Perf. score & Rob. score & Avg. score \\ [0.5ex] 
 \hline \hline
TVLQR & 0.504 & 0.607 & 0,556\\
\hline
iLQR MPC stab. & 0.345 & 0.343 & 0.344 \\
\hline 
\rowcolor{LightRed} MC-PILCO & 0.316 & 0.25 & 0.283 \\
 \hline
iLQR Riccati & 0.396 & 0.138 & 0.267 \\
\hline

\end{tabular}
\caption{Acrobot scores comparison.}
\label{tab:acrobot_table}
\end{table}
\end{center}

\section*{Acknowledgements}
Alberto Dalla Libera was supported by PNRR research activities of the consortium iNEST (Interconnected North-Est Innovation Ecosystem) funded by the European Union Next GenerationEU (Piano Nazionale di Ripresa e Resilienza (PNRR) – Missione 4 Componente 2, Investimento 1.5 – D.D. 1058  23/06/2022, ECS\_00000043). This manuscript reflects only the Authors’ views and opinions, neither the European Union nor the European Commission can be considered responsible for them.

\bibliographystyle{ieeetr}
\bibliography{ref}

\begin{thebibliography}{10}

\bibitem{sutton2018reinforcement_learning}
R.~S. Sutton and A.~G. Barto, {\em Reinforcement learning: An introduction}.
\newblock MIT press, 2018.

\bibitem{amadio2021mc_pilco}
F.~Amadio, A.~Dalla~Libera, R.~Antonello, D.~Nikovski, R.~Carli, and D.~Romeres, ``Model-based policy search using monte carlo gradient estimation with real systems application,'' {\em IEEE Transactions on Robotics}, vol.~38, no.~6, pp.~3879--3898, 2022.

\bibitem{ijcai2024p1043}
F.~Wiebe, N.~Turcato, A.~Dalla~Libera, C.~Zhang, T.~Vincent, S.~Vyas, G.~Giacomuzzo, R.~Carli, D.~Romeres, A.~Sathuluri, M.~Zimmermann, B.~Belousov, J.~Peters, F.~Kirchner, and S.~Kumar, ``Reinforcement learning for athletic intelligence: Lessons from the 1st “ai olympics with realaigym” competition,'' in {\em Proceedings of the Thirty-Third International Joint Conference on Artificial Intelligence, {IJCAI-24}} (K.~Larson, ed.), pp.~8833--8837, International Joint Conferences on Artificial Intelligence Organization, 8 2024.
\newblock Demo Track.

\bibitem{amadio2023mcpilco_raw_meas}
F.~Amadio, A.~Dalla~Libera, D.~Nikovski, R.~Carli, and D.~Romeres, ``Learning control from raw position measurements,'' in {\em 2023 American Control Conference (ACC)}, pp.~2171--2178, IEEE, 2023.

\bibitem{mcpilco_tossing}
N.~Turcato, A.~D. Libera, G.~Giacomuzzo, and R.~Carli, ``Teaching a robot to toss arbitrary objects with model-based reinforcement learning,'' in {\em 2023 9th International Conference on Control, Decision and Information Technologies (CoDIT)}, pp.~1126--1131, 2023.

\bibitem{10375556}
F.~Wiebe, S.~Kumar, L.~J. Shala, S.~Vyas, M.~Javadi, and F.~Kirchner, ``Open source dual-purpose acrobot and pendubot platform: Benchmarking control algorithms for underactuated robotics,'' {\em IEEE Robotics \& Automation Magazine}, vol.~31, no.~2, pp.~113--124, 2024.

\bibitem{rasmussen2003gps_for_ml}
C.~E. Rasmussen, ``Gaussian processes in machine learning,'' in {\em Summer school on machine learning}, pp.~63--71, Springer, 2003.

\bibitem{kernel_methods_and_gp_control_systems_magazine}
A.~Carè, R.~Carli, A.~D. Libera, D.~Romeres, and G.~Pillonetto, ``Kernel methods and gaussian processes for system identification and control: A road map on regularized kernel-based learning for control,'' {\em IEEE Control Systems Magazine}, vol.~43, no.~5, pp.~69--110, 2023.

\bibitem{caflisch1998monte_carlo_sampling_ref}
R.~E. Caflisch, ``Monte carlo and quasi-monte carlo methods,'' {\em Acta numerica}, vol.~7, pp.~1--49, 1998.

\bibitem{bottou2010large_scale_learning_sgd}
L.~Bottou, ``Large-scale machine learning with stochastic gradient descent,'' in {\em Proc of COMPSTAT'2010}, pp.~177--186, Springer, 2010.

\bibitem{kingma2013reparametrization_trick}
D.~P. Kingma and M.~Welling, ``Auto-encoding variational bayes,'' {\em arXiv preprint arXiv:1312.6114}, 2013.

\bibitem{srivastava2014dropout}
N.~Srivastava, G.~Hinton, A.~Krizhevsky, I.~Sutskever, and R.~Salakhutdinov, ``Dropout: a simple way to prevent neural networks from overfitting,'' {\em JMLR}, vol.~15, no.~1, pp.~1929--1958, 2014.

\bibitem{amadio2023learning}
F.~Amadio, A.~D. Libera, D.~Nikovski, R.~Carli, and D.~Romeres, ``Learning control from raw position measurements,'' 2023.

\bibitem{JMLR:v6:quinonero-candela05a}
J.~Qui{{\~n}}onero-Candela and C.~E. Rasmussen, ``A unifying view of sparse approximate gaussian process regression,'' {\em Journal of Machine Learning Research}, vol.~6, no.~65, pp.~1939--1959, 2005.

\bibitem{paszke2017automatic_diff_pytorch}
A.~Paszke, S.~Gross, S.~Chintala, G.~Chanan, E.~Yang, Z.~DeVito, Z.~Lin, A.~Desmaison, L.~Antiga, and A.~Lerer, ``Automatic differentiation in pytorch,'' 2017.

\end{thebibliography}

\end{document}